\documentclass{article}

\usepackage[english]{babel}
\usepackage{amsfonts}

\usepackage[letterpaper,top=2cm,bottom=2cm,left=3cm,right=3cm,marginparwidth=1.75cm]{geometry}

\usepackage[utf8]{inputenc}
\usepackage{csquotes}
\usepackage{amsmath}
\usepackage{graphicx}
\usepackage{pmboxdraw}
\usepackage{multirow}
\usepackage{comment}
\usepackage[colorlinks=true, allcolors=blue]{hyperref}

\newcommand{\spiece}{▁}

\usepackage[
backend=biber,
style=alphabetic,
sorting=ynt
]{biblatex}

\title{Training and Evaluation of a Multilingual Tokenizer for GPT-SW3}
\author{
Felix Stollenwerk \\ AI Sweden}
\date{}

\begin{document}

\maketitle

\begin{abstract}
This paper provides a detailed discussion of the multilingual tokenizer used for GPT-SW3. It was trained on the Nordic Pile using the SentencePiece library and the BPE algorithm. We outline the tokenizer's most important features and share details on its learned vocabulary. In addition, we systematically analyze the properties and evaluate the performance of the tokenizer with regard to the different languages present in the data.
\end{abstract}

\section{Introduction \label{sec:introduction}}

\begin{figure}[ht]
    \centering
    \includegraphics[scale=0.38]{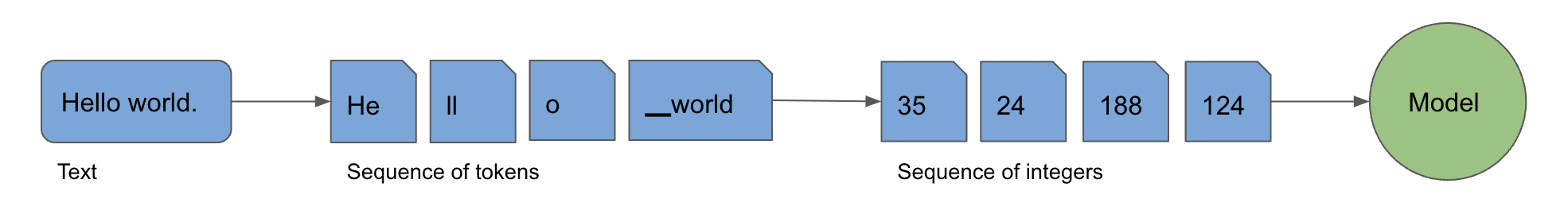}
    \caption{Text preprocessing for language models (simplified). The first step is referred to as tokenization, although sometimes both the first and second step are embraced by the same term. Note that the character ▁ which appears in the above example represents whitespace (more on this in Sec.~\ref{sec:tokenizer_features}).}
    \label{fig:tokenization_example}
\end{figure}

Generative language models are pre-trained on large amounts of raw text data. Virtually all language model architectures require the text data to be tokenized, which means that a text string is split into a sequence of tokens and subsequently mapped to a sequence of integers, see Fig.~\ref{fig:tokenization_example}.

Modern subword tokenizers are designed such that frequently used words are not decomposed while rare words are split into meaningful tokens.   
Two of the most commonly used subword tokenizer training algorithms (see App.~\ref{app:literature review} for a literature overview) are Byte-Pair Encoding (BPE) \cite{sennrich2016neural} and Unigram \cite{kudo2018subword}. We chose a BPE tokenizer for GPT-SW3 \cite{gptsw3} as it is well-supported by NeMo Megatron \cite{shoeybi2020megatronlm}, which is the framework we used for model training.

A BPE tokenizer is defined by its vocabulary, which is an ordered list of tokens. When it is applied to a text string, the tokenizer goes through its vocabulary from top to bottom and merges the characters of a given token whenever it appears in the data. The vocabulary tokens themselves are determined by the vocabulary size and the BPE algorithm. The vocabulary size is a hyperparameter that needs to be specified beforehand. It typically lies between 30000 and 50000, but larger vocabulary sizes have also been used (see App.~\ref{app:literature review}). 
The BPE algorithm essentially works like this\footnote{see \url{https://huggingface.co/docs/transformers/tokenizer_summary##bytepair-encoding-bpe} for more details}: First, the training data is split into characters, the initial tokens. Afterwards, the algorithm goes through the whole list of tokens iteratively. At each iteration, it determines the pair of tokens that appears together the most often. The token pair at hand is then added to the vocabulary and all its occurrences are merged. The process is repeated until the vocabulary has reached its predetermined size. 
In order for tokenization to work particularly well for the data the model is trained on, a custom tokenizer is usually trained on the very same data.

This document is structured as follows: Sec.~\ref{sec:data} describes the data which is used for the training of the tokenizer. In Sec.~\ref{sec:tokenizer_features}, we discuss specific features of the tokenizer and how it is trained. Subsequently, in Sec.~\ref{sec:tokenizer_evaluation}, we introduce a way to evaluate the tokenizer. After those theoretical parts, results are presented in Sec.~\ref{sec:tokenizer_results}. Finally, in Sec.~\ref{sec:multilingualism}, we extend our analysis by comparing the tokenizer with monolingual tokenizers and by studying the impact of the vocabulary size. Conclusions can be found in Sec.~\ref{sec:conclusions}.

\section{Data \label{sec:data}}

The tokenizer, just like the model, is trained on data from the Nordic Pile \cite{öhman2023nordic}. This dataset contains different subsets characterized by category and language. As discussed in detail in \cite{gptsw3}, those subsets are weighted before they are used for tokenizer and model training, with the same weights being used in both cases.
However, in contrast to the training of the model, it is prohibitively expensive and unnecessary to train the tokenizer on the whole dataset. Instead, we sample $1\%$ of the original data while preserving the weighting of the different category and language subsets. 
In addition, we create a few monolingual datasets for the purpose of evaluating the tokenizer (more on this in Sec.~\ref{sec:tokenizer_evaluation}). More specifically, we sample another multilingual dataset, containing $0.2\%$ of the original data while again preserving the weighting of the category subsets. Then, we split this dataset into its monolingual parts, one for each of the languages present in the Nordic Pile. 
An overview of all the datasets we use is given in Tab.~\ref{tab:datasets}.
\begin{table}[ht]
    \centering
    \begin{tabular}{lccc} \hline
    \textbf{purpose} & \textbf{sampling fraction} & \textbf{language} & \textbf{no. of datasets}  \\ \hline
    training & $1\%$ & mulitlingual & 1 \\
    evaluation & $0.2\%$ & monolingual & 6 \\ \hline
    \end{tabular}
    \caption{Overview of the datasets we use.}
    \label{tab:datasets}
\end{table}

\section{Training \label{sec:tokenizer_features}}

The tokenizer is trained using SentencePiece \cite{kudo-richardson-2018-sentencepiece}. While this library supports the BPE algorithm, it provides a few features on top. Most notably, whitespace is mapped to a special \texttt{\spiece} unicode character (\texttt{U+2581}) and SentencePiece provides lossless (i.e.~reversible) tokenization without the need to pre-tokenize. 
In this section, we will describe the most important features\footnote{Features that we considered but ended up not using are briefly discussed in App.~\ref{app:tokenizer_other_features}.} of our tokenizer. Technical details regarding their implementation are discussed in App.~\ref{app:tokenizer_sentencepiece}.

\paragraph{F1) Special Tokens}
SentencePiece allows for the use of four special tokens, namely for padding, to mark the begin and end of a sequence (i.e. a document) and to handle unknown tokens. In our case, they represent the first four tokens in our tokenizer's vocabulary. The order and choice of symbols are largely determined by NeMo Megatron and can be found in App.~\ref{app:tokenizer_sentencepiece}. Note that they are called special tokens as they are added or created during data preprocessing or tokenization, but do not appear in the text data itself. 

\paragraph{F2) Split Digits}
We arrange for our tokenizer to automatically split digits into single character tokens. As an example, \texttt{123.4} is tokenized to \texttt{[1~2~3~.~4]}. Without this special rule, it might be tokenized to something like \texttt{[123~.~4]} instead. Splitting digits like this arguably improves the handling of numbers and mathematical capabilities of a model. 

\paragraph{F3) Dummy Prefix}
Tokenization with SentencePiece works such that whitespace is absorbed in the first token of a word. By default, this leads to the somewhat counterintuitive situation that a word's tokenization depends on whether it is located at the beginning of a document or not\footnote{see \url{https://discuss.huggingface.co/t/bpe-tokenizers-and-spaces-before-words/475} for more details}. For instance, the word \texttt{Swedish} may be tokenized as \texttt{[Swed ish]} (beginning of document) or \texttt{[\_Swed ish]} (elsewhere). SentencePiece allows to enforce consistent representations by adding a so-called dummy prefix to the beginning of text. We make use of this feature, and thereby follow e.g. \cite{black2022gptneox20b}.   

\paragraph{F4) Whitespace Concatenation}
Some text data contain significant amounts of whitespace, like for instance code with indentation. In order for the tokenizer to decompose such text into fewer tokens, one may add consecutive whitespace tokens to the vocabulary to ensure that whitespace is always concatenated. The method was introduced in \cite{black2022gptneox20b} using consecutive whitespace tokens of length between 2 and 24, and we adopt this approach.  

\paragraph{F5) Code Tokens}
The code data we use contains four different programming languages (see \cite{öhman2023nordic}). In order to help the model differentiate between them, a specific code token which indicates the language is added to the beginning of each code document. Since we want the tokenizer to preserve the meaning of those code tokens without breaking them down to subtokens, we add them to the vocabulary as well.  

\paragraph{F6) Byte Fallback and Character Coverage}
During tokenizer training, all single (unicode) characters present in the training data are added to the vocabulary. While this often covers a wide range of different characters, it can't be excluded that the tokenizer will encounter previously unseen characters when it is applied to new data.
The default behaviour of a tokenizer in order to deal with these cases is to assign a special unknown token (see feature F1) to words containing a character that is not present in the vocabulary. However, the meaning of the word in question is lost in that case.
An alternative way to handle the issue without the need for unknown tokens is a feature called byte fallback. It entails that all 256 UTF-8 byte code units are added to the vocabulary. Any unknown unicode character may then be decomposed into byte code units\footnote{each unicode character is mapped to 1-4 byte code units, see e.g. \url{https://en.wikipedia.org/wiki/UTF-8}}. 
This leads to a unique, arguably more meaningful representation of a word in question. 
Another feature, related to byte fallback, is character coverage. Instead of adding all (unicode) characters encountered in the training data to the vocabulary, as outlined above, one may opt to restrict the number of characters and treat some rare characters using byte fallback.
While this feature is especially useful for character-rich languages (e.g. Chinese), it may also generally be a good idea to save some space in the vocabulary for regular, multi-character tokens instead of extremely rare and often not very meaningful unicode characters. We opt to use byte fallback with a character coverage of $0.9999$.

\paragraph{F7) Vocabulary Size}
Choosing the vocabulary size is a trade-off. While a larger vocabulary leads to a better representation of text that the model performance may benefit from (cf.~Sec.~\ref{sec:introduction}), it also increases the size of the embedding matrix and hence the computational cost at model training and inference time.  
We use 
\begin{equation}
    \text{vocabulary size} = 64000
    \label{eq:vocabulary_size}
\end{equation}
This choice is motivated mainly by the fact that most of the LLMs in the literature, previously trained on a single language (i.e. English), use a vocabulary size of around 50000 (see App.~\ref{app:literature review}). Since our tokenizer and model need to accommodate multiple (although related) languages, we opt for a slightly larger vocabulary size. The effect of the vocabulary size on the performance of the tokenizer will be studied in Sec.~\ref{sec:multilingualism}.

\section{Evaluation \label{sec:tokenizer_evaluation}}

As briefly discussed in Sec.~\ref{sec:introduction}, a good tokenizer splits text in such a way that the resulting sequence of tokens is a meaningful representation which the model will be able to learn from efficiently. While this definition of quality is rather vague and not directly measurable, a meaningful representation will typically include many whole words and few subword tokens. Two useful metrics that can be used as a proxy to quantify the quality of a tokenizer with respect to a given dataset are \textit{fertility} $f$ and \textit{proportion of continued words} $p$ \cite{acs, rust-etal-2021-good}.
Fertility is the average number of subwords that a word is decomposed into. In our case, this number is equivalent to the total number of tokens, divided by the number of tokens that start with \spiece. Special care needs to be taken of punctuation, as commas and full stops are one-character tokens that can't be split, yet do not contain \spiece. We decide to exclude them from the fertility computation as we think that this approach leads to a cleaner assessment. The proportion of continued words is the number of words that are split, divided by the total number of words. 
For the sake of consistency, we again exclude punctuation.
In contrast to fertility, the proportion of continued words does not capture \textit{how often} a word is split, but rather only \textit{whether} a word is split. Hence, the two metrics are somewhat complementary. They assume values $f \geq 1$ and $0 \leq p \leq 1$, with lower values indicating that a tokenizer is well-suited for the dataset it is evaluated on.
In Sec.~\ref{sec:tokenizer_results}, we will apply our multilingual tokenizer (cf.~Sec.~\ref{sec:tokenizer_features}) to the different evaluation datasets (cf.~Sec.~\ref{sec:data}) and compute the two evaluation metrics. The results will give us an indication of how well the tokenizer works for the different languages present in our dataset.

\section{Results \label{sec:tokenizer_results}}

\subsection{Vocabulary}

\begin{figure}[ht]
    \centering
    \includegraphics[scale=0.7]{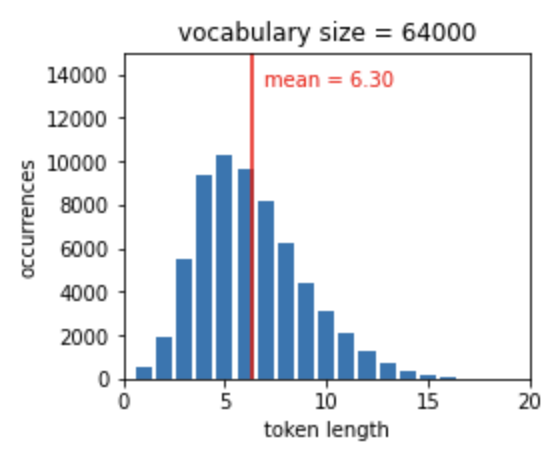}
    \caption{Histogram of the token length in the vocabulary of the tokenizer. Note that there are 554 one-character tokens (cf.~Tab.~\ref{tab:tokenizer_vocabulary}). The longest regular tokens contain 16 characters.}
    \label{fig:tokenizer_result_vocabulary}
\end{figure}

In Tab.~\ref{tab:tokenizer_vocabulary}, we list the different groups of tokens in our tokenizer's vocabulary.
\begin{table}[ht]
    \centering
    \begin{tabular}{lrll} \hline
    \textbf{token numbers} & \textbf{count} & \textbf{group} & \textbf{tokens} \\ \hline
    1-4 & 4 & special tokens & \textlangle pad\textrangle, \textlangle unk\textrangle, \textlangle s\textrangle, \textlangle $|$endoftext$|$\textrangle \\
    5-8 & 4 & code tokens & \textlangle $|$javascript$|$\textrangle, \textlangle $|$python$|$\textrangle, $\ldots$ \\
    9-264 & 256 & byte fallback tokens & \textlangle 0x00\textrangle, \textlangle 0x01\textrangle, $\ldots$ \\
    265-63423 & 63159 & regular tokens & er, en, $\ldots$ \\
    63424-63977 & 554 & one-character tokens & e, n, $\ldots$ \\
    63978-64000 & 23 & consecutive whitespace tokens & \spiece\spiece, \spiece\spiece\spiece, $\ldots$ \vspace{0.3mm} \\ \hline
    \end{tabular}
    \caption{Overview of the tokenizer vocabulary.}
    \label{tab:tokenizer_vocabulary}
\end{table}
Note that the first and last groups, namely the special tokens, code tokens, byte fallback tokens and consecutive whitespace tokens, were explicitly added to the vocabulary. In contrast, the regular tokens and one-character tokens are determined by the data in conjunction with the BPE algorithm.
Fig.~\ref{fig:tokenizer_result_vocabulary} shows the distribution of the token length in the vocabulary.

\subsection{Examples}

\begin{figure}[ht]
    \centering
    \includegraphics[scale=0.46]{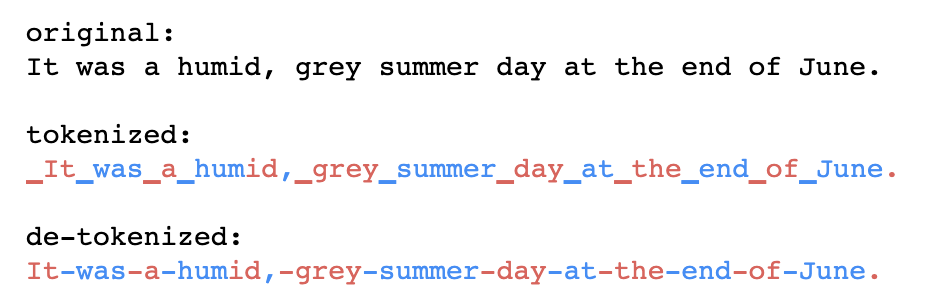}
    \includegraphics[scale=0.46]{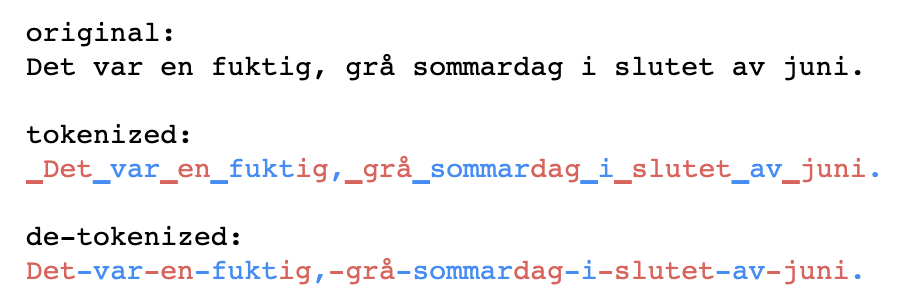}
    \caption{Tokenization example using English (left) and Swedish (right) text.}
    \label{fig:tokenization_example_en_sv}
\end{figure}

\begin{figure}[ht]
    \centering
    \includegraphics[scale=0.43]{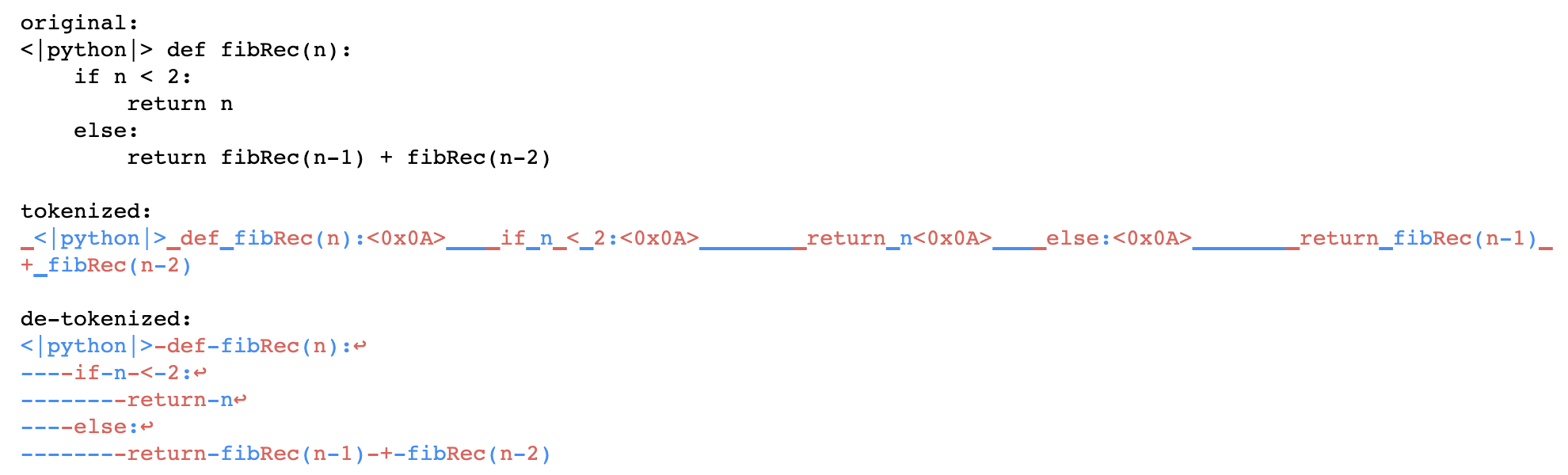}
    \caption{Tokenization example using python code. Note that line breaks are explicitly shown in the detokenized text.}
    \label{fig:tokenization_example_cd}
\end{figure}

In this section, we show a few examples of how text is processed by our tokenizer. In particular, Fig.~\ref{fig:tokenization_example_en_sv} contains an example sentence in English and Swedish, while Fig.~\ref{fig:tokenization_example_cd} corresponds to a code example. 
Each example displays the original text, the tokenized text and the detokenized text.
Technically speaking, the tokenized text is a list of strings. However, in the figures, we display it as a single string to emphasize the mapping from the original text, with the tokens being highlighted alternately in red and blue. 
The detokenized text is a string like the original text. In fact, due to the reversibility of the tokenization process (see Sec.~\ref{sec:tokenizer_features}), they are exactly the same. However, in the figures, we again highlight the substrings that correspond to the tokens. In order to do so, whitespace is represented by hyphens.    

Apart from the aforementioned reversibility and the use of the 
\spiece~character typical for SentencePiece, the examples illustrate some of the specific features (see Sec.~\ref{sec:tokenizer_features}) we employed for our tokenizer. Firstly, all the examples reveal that the first token of a document always starts with a dummy prefix (F3). Secondly, the code example (Fig.~\ref{fig:tokenization_example_cd}) concatenates whitespace (F4), preserves the python code token (F5), and uses the byte fallback feature (F6) for line breaks, which are mapped to \textlangle0x0A\textrangle.

\subsection{Evaluation}

\begin{figure}[ht]
    \centering
    \includegraphics[scale=0.5]{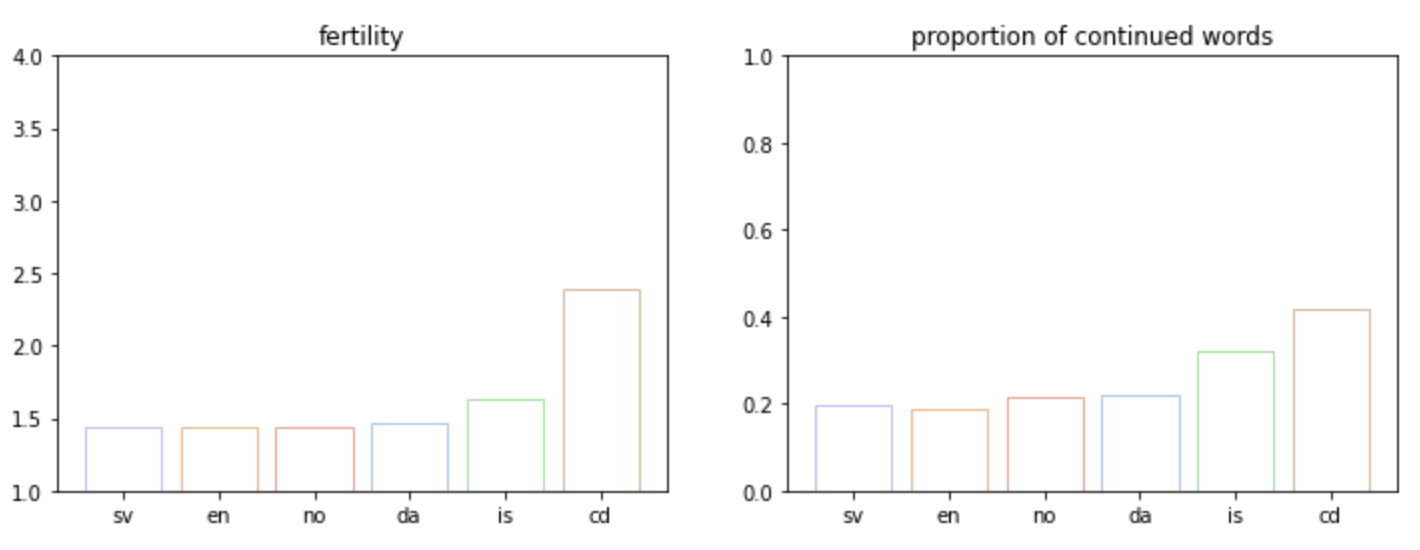}
    \caption{Fertility (left) and proportion of continued words (right) evaluated using our multilingual tokenizer on evalution datasets of different languages (horizontal axis).}
    \label{fig:tokenizer_result_evaluation}
\end{figure}

We apply the tokenizer (see Sec.~\ref{sec:tokenizer_features}) to the evalution datasets of different languages (see Sec.~\ref{sec:data}) and compute the evaluation metrics fertility and proportion of continued words (see Sec.~\ref{sec:tokenizer_evaluation}) separately for each of the languages. The results are shown in Fig.~\ref{fig:tokenizer_result_evaluation}.

We find that both evaluation metrics assume similar values for Swedish, English, Norwegian and Danish. Icelandic and especially code stand out with significantly higher values. This is likely due to the fact that these languages are underrepresentated in the training data, cf.~Sec.~\ref{sec:data}. However, we can't exclude that there are other, language-specific factors which contribute to the observed effect\footnote{For instance, code often contains custom variable names and thus may naturally contain a larger amount of unique words than the other languages. These will probably be decomposed into many subword tokens, which may contribute to the exceptionally high value of fertility. However, further investigation would be needed to scrutinize this hypothesis.}.
We conclude that the amount to which Swedish, English, Norwegian and Danish are represented in our tokenizer's vocabulary is approximately the same. In contrast, Icelandic and even more so code are less accounted for by the tokenizer. In consideration of the results from \cite{rust-etal-2021-good}, we have to assume that the performance of the language model with regard to those languages will be affected negatively.   

\section{Analysis \label{sec:multilingualism}}

In this section, we extend the analysis of our multilingual tokenizer in two ways. First, we train monolingual tokenizers for the different languages on the respective monolingual fractions of the training dataset, using the same vocabulary size as before (see Eq.~({\ref{eq:vocabulary_size}})). Their properties and performance are compared to the multilingual tokenizer in Sec.~\ref{subsec:multilingualism_fixed}. 
Afterwards, in Sec.~\ref{subsec:multilingualism_varied}, we vary the vocabulary size of the multilingual tokenizer and study the impact this has on the tokenizer's performance.

\subsection{Monolingual tokenizers \label{subsec:multilingualism_fixed}} 

We compare the multilingual and monolingual tokenizer for each language in two separate ways: by comparing the performance in terms of fertility and proportion of continued words, and by determining the vocabulary overlap.

\begin{figure}[ht]
    \centering
    \includegraphics[scale=0.5]{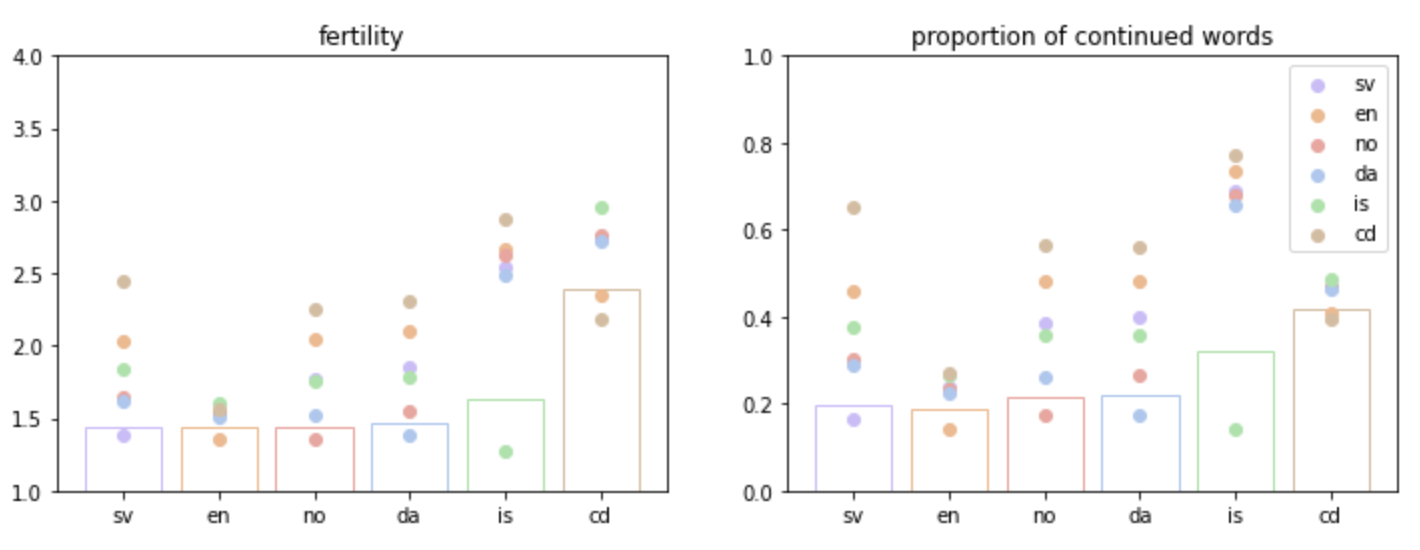}
    \caption{Fertility (left) and proportion of continued words (right) evaluated using our multilingual tokenizer (bars) and monolingual tokenizers (dots) on evalution datasets of different languages (horizontal axis). Compare to Fig.~\ref{fig:tokenizer_result_evaluation}.}
    \label{fig:multi_evaluation_fixed}
\end{figure}

\paragraph{Performance} We apply and evaluate each monolingual tokenizer on each of the evaluation datasets. Since we have 6 different languages, this gives us 36 combinations of monolingual tokenizers and monolingual data. 
In Fig.~\ref{fig:multi_evaluation_fixed}, 
we show the results for fertility and proportion of continued words, together with the results for the multilingual tokenizer that we obtained earlier.
We observe the following patterns:
\begin{itemize}
    \item For every language, the corresponding monolingual tokenizer always performs best, followed by the multilingual tokenizer (with code data being an exception). The gap between them is relatively small, except for Icelandic.
    \item All the monolingual tokenizers work reasonably well for English text. The reason might be that English is widely used both by Scandinavian native speakers and in code.
    \item The Swedish, Norwegian and Danish tokenizers work relatively well on other languages of the same group. This is not unexpected and most likely due to the similarity of the languages.
    \item The aforementioned patterns are very similar for the two evaluation metrics, fertility and proportion of continued words.
\end{itemize}

\paragraph{Vocabulary overlap} 

\begin{figure}[ht]
    \centering
    \includegraphics[scale=0.5]{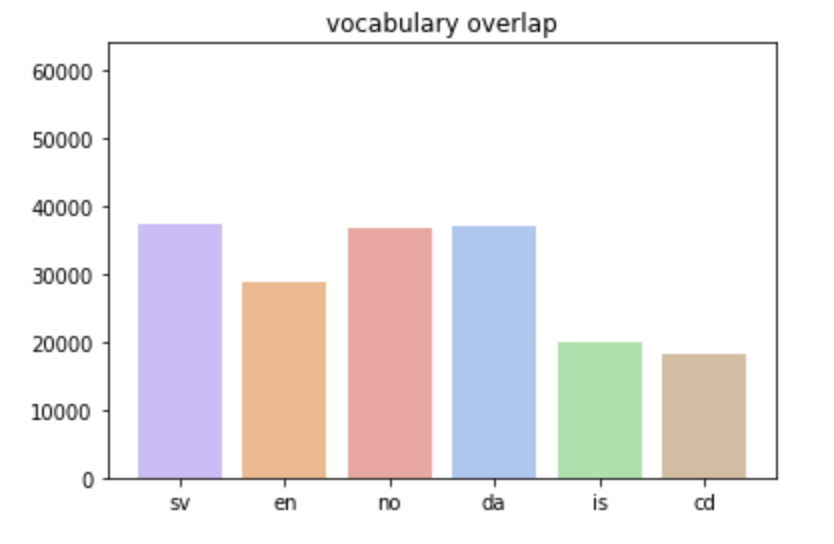}
    \caption{Vocabulary overlap of the multilingual tokenizer with the respective monolingual tokenizers.}
    \label{fig:multi_vocabulary_overlap_fixed}
\end{figure}

In Fig.~\ref{fig:multi_vocabulary_overlap_fixed},  
we show the vocabulary overlap between the monolingual tokenizers and the multilingual tokenizer.
Our findings are:
\begin{itemize}
    \item The multilingual tokenizer has the largest vocabulary overlap  with the monolingual tokenizers trained on Swedish, Norwegian and Danish text. Among those languages, the differences are minimal.
    \item The vocabulary overlap with respect to English is somewhat smaller. A potential cause for this is that---despite English being the second biggest language in our corpus---the aforementioned Scandinavian languages profit from each other because of the high similarity and overlap in words.
    \item Icelandic and code are the least present in the multilingual tokenizer vocabulary. This is presumably mainly due to their underrepresentation in the data.
\end{itemize}

\subsection{Impact of the vocabulary size \label{subsec:multilingualism_varied}}

In this section, we vary the vocabulary size of the multilingual tokenizer, using the values 10000, 20000, 30000, 40000, 51200 and 64000. 
We then study the impact this has on the performance and the vocabulary overlap with the monolingual tokenizers.

\paragraph{Performance}

\begin{figure}[ht]
    \centering
    \includegraphics[scale=0.5]{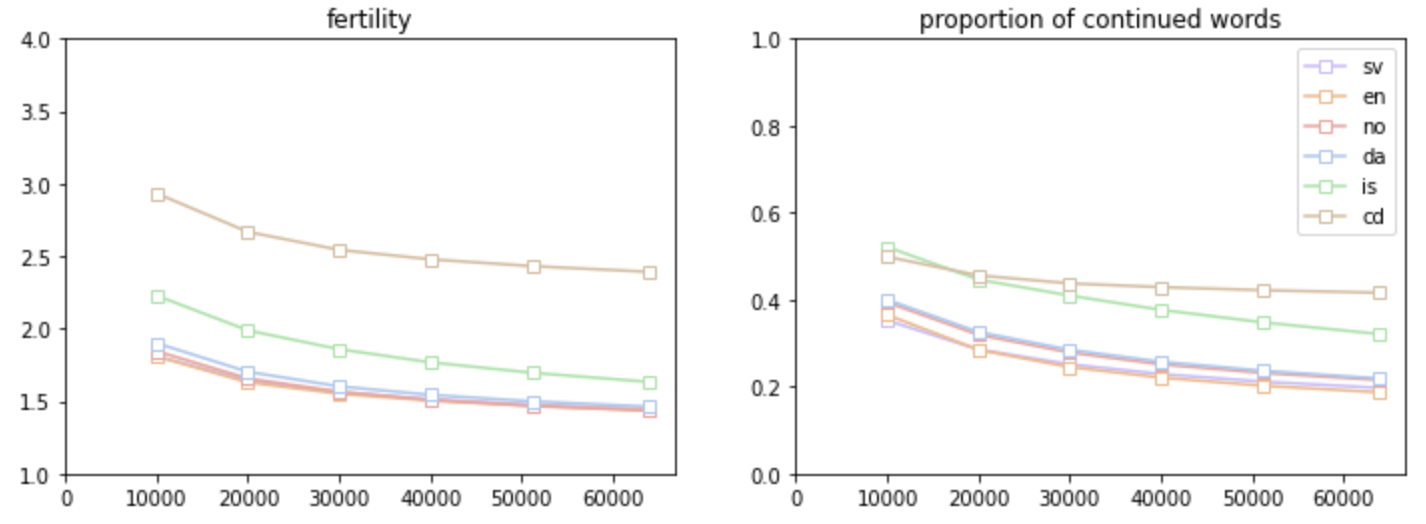}
    \caption{Fertility (left) and proportion of continued words (right) evaluated using our multilingual tokenizer on different languages (colors) with different vocabulary sizes (horizontal axis). Note that the values at vocabulary size = 64000 correspond to Fig.~\ref{fig:tokenizer_result_evaluation}.}
    \label{fig:multi_evaluation_varied}
\end{figure}

Fig.~\ref{fig:multi_evaluation_varied} shows the evaluation metrics for the different languages as functions of the multilingual tokenizer's vocabulary size.
We observe, as expected, that larger vocabulary size leads to better tokenizer performance. Moreover, the curves seem to flatten as the vocabulary size increases, which is an additional reason for our choice in Eq.~(\ref{eq:vocabulary_size}). 

\begin{figure}[ht]
    \centering
    \includegraphics[scale=0.5]{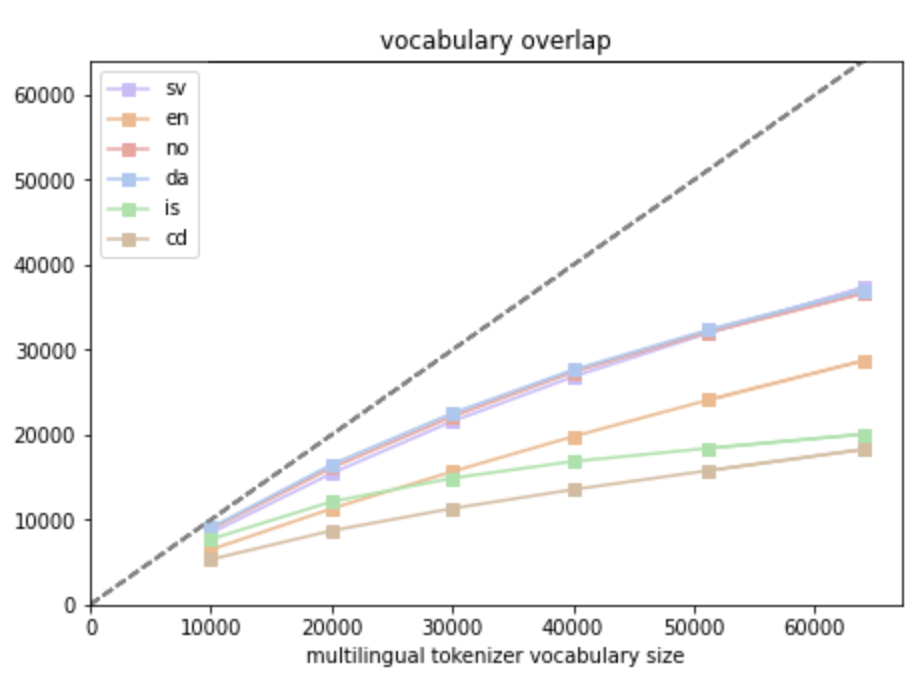}
    \caption{Vocabulary overlap of the multilingual tokenizer with the respective monolingual tokenizers, as a function of the multilingual tokenizer's vocabulary size. Note that the values at vocabulary size = 64000 correspond to Fig.~\ref{fig:multi_vocabulary_overlap_fixed}.}
    \label{fig:multi_vocabulary_overlap_varied}
\end{figure}

\paragraph{Vocabulary overlap} 
Fig.~\ref{fig:multi_vocabulary_overlap_varied} shows the vocabulary overlap between the multilingual tokenizer and the monolingual tokenizers, similar to Fig.~\ref{fig:multi_vocabulary_overlap_fixed}, but as a function of the multilingual vocabulary size.
We find that larger vocabulary sizes of the multilingual tokenizer lead to better representations of the languages in the vocabulary. For Icelandic, we observe a stronger flattening of the curve compared to the other languages. This means that Icelandic profits less from a larger multilingual vocabulary size than the other languages. 

\section{Conclusions \label{sec:conclusions}}

We have trained and evaluated a multilingual tokenizer for the GPT-SW3 model on the Nordic Pile. We find that the multilingual tokenizer has very similar features and performance regarding Swedish, Norwegian and Danish. The performance with respect to English is similarly good, although the underlying mechanisms causing this seem to be different than for the Scandinavian languages. In particular, English is less well represented in the multilingual tokenizer's vocabulary. Moreover, the multilingual tokenizer turns out to be less suited for Icelandic and code. In the light of \cite{rust-etal-2021-good}, the results are an important step to disentangle the effect of the tokenizer and the effect of the training data on the language model's performance with respect to the different languages.

\printbibliography

\appendix

\section{Literature Review \label{app:literature review}}

\begin{table}[ht]
    \centering
    \begin{tabular}{llllll} \hline
    \textbf{date} & \textbf{paper} & \textbf{tokenizer} & \textbf{languages} & \textbf{dataset size} & \textbf{vocabulary size}\\ \hline
    2019-01 & XLM & BPE & 15 & unknown & 100k \\
    2019-02 & GPT-2 & BPE & English & 40GB & 50k (+257) \\ 
    2019-10 & T5 & Unigram/SP & English & 750GB & 32k \\
    2020-05 & GPT-3 & BPE & English & 300B tokens & 50k (+257) \\
    2021-08 & Jurassic-1 & Unigram/SP & English & 300B tokens & 256k \\
    2021-12 & Gopher & BPE/SP & English ($>99\%$) & 300B tokens & 32k \\
    2022-03 & Chinchilla & BPE/SP & English & 270B tokens & 32k \\
    2022-04 & GPT-NeoX-20B & BPE & English & 826G & 50k (+257) \\
    2022-04 & PaLM & Unigram/SP & 100 & 780B tokens & 256k \\
    \hline
    \end{tabular}
    \caption{Informal and non-exhaustive overview of language model tokenizers and their most important features. SP stands for SentencePiece.}
    \label{tab:literature_overview}
\end{table}

\section{Tokenizer features that we did not use \label{app:tokenizer_other_features}}

Our tokenizer does not employ any kind of unicode normalization, as the data from the Nordic Pile is already normalized (using NFC). Another feature that we considered but ended up not using is the imposition of a minimum frequency. This means that only tokens that occur more often than a certain threshold are included in the vocabulary. However, this leads to a vocabulary size which is not determined beforehand and most likely not divisible by 128 (which is a constraint in NeMo Megatron). Setting the vocabulary size directly as in Eq.~(\ref{eq:vocabulary_size}) avoids those issues and effectively leads to a minimum frequency requirement as well (albeit less strict). 

\section{Tokenizer feature implementation \label{app:tokenizer_sentencepiece}}

Most of the features mentioned in Sec.~\ref{sec:tokenizer_features} are readily available in SentencePiece. They can be used by simply specifying certain parameters in the \texttt{train} function of SentencePiece's Python API. 
In Tab.~\ref{tab:tokenizer_features}, we list the parameters and values we used, together with the features they correspond to. 
\begin{table}[ht]
    \centering
    \begin{tabular}{lll} \hline
    \textbf{feature} & \textbf{parameter} & \textbf{value} \\ \hline
    \multirow{8}{*}{F1} & \texttt{pad\_id} & 0 \\
    & \texttt{unk\_id} & 1 \\
    & \texttt{bos\_id} & 2 \\
    & \texttt{eos\_id} & 3 \\ \cline{2-3}
    & \texttt{pad\_piece} & \textlangle pad\textrangle \\
    & \texttt{unk\_piece} & \textlangle unk\textrangle \\
    & \texttt{bos\_piece} & \textlangle s\textrangle \\
    & \texttt{eos\_piece} & \textlangle $|$endoftext$|$\textrangle \\ \hline
    F2 & \texttt{split\_digits} & True \\ \hline
    F3 & \texttt{add\_dummy\_prefix} & True \\ \hline
    F4 & --- & --- \\ \hline
    F5 & \texttt{user\_defined\_symbols} & \textlangle $|$javascript$|$\textrangle, \textlangle $|$python$|$\textrangle, \textlangle $|$sql$|$\textrangle, \textlangle $|$shell$|$\textrangle \\ \hline
    \multirow{2}{*}{F6} & \texttt{byte\_fallback} & True \\
    & \texttt{character\_coverage} & 0.9999 \\ \hline
    F7 & \texttt{vocabulary\_size} & 64000 \\
    \hline
    \end{tabular}
    \caption{Overview of parameters and values used in the SentencePiece \texttt{train} function. The first column refers to the features described in Sec.~\ref{sec:tokenizer_features}.}
    \label{tab:tokenizer_features}
\end{table}

An exception to the above is whitespace concatenation (F4), which we had to implement ourselves. We did so by manually replacing the last 23 regular tokens in the vocabulary (after training) by consecutive whitespace tokens of length 2-24, i.e. \spiece\spiece (= 2x \texttt{U+2581}), \spiece\spiece\spiece (= 3x \texttt{U+2581}) etc. Note that this is also reflected in Tab.~\ref{tab:tokenizer_vocabulary}.

\end{document}